%% file: robust_blind.tex
\documentclass[twoside,11pt]{article}
\pdfoutput=1
\usepackage{jmlr2e}
\usepackage{amsmath,graphicx,amsfonts,amssymb,amscd,amsmath}
\usepackage{times}
\usepackage{epsfig}
\usepackage{graphicx}
\usepackage{amsmath}
\usepackage{amssymb}
\usepackage{algorithm}
\usepackage{algorithmic}
\usepackage{color,paralist}
\usepackage{comment}
\usepackage{subcaption}
\usepackage{caption}

\newcommand{\R}{\mathbb{R}}

\newcommand{\loss}{\mathcal{L}}

\newcommand{\KL}{KL}
\usepackage{array}
\usepackage{booktabs}
\usepackage{caption}
\usepackage{wrapfig}
\usepackage{hyperref}

\graphicspath{{figs/}}

\usepackage{lineno}

\newlength\myindent
\makeatletter
\let\@oldmaketitle\@maketitle
\renewcommand{\@maketitle}{\@oldmaketitle
	\center
	\includegraphics[height=1.3in]{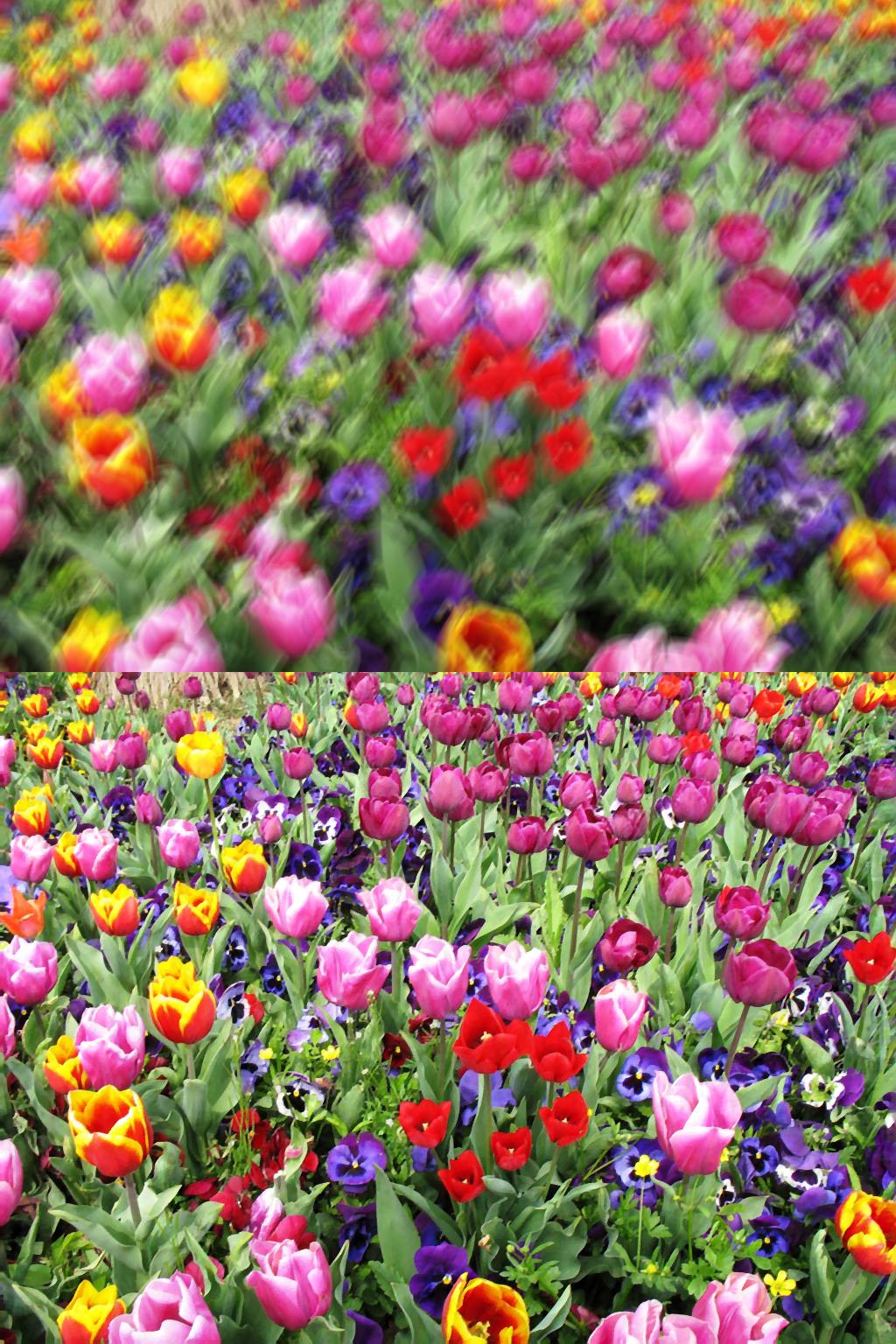}
	\includegraphics[height=1.3in]{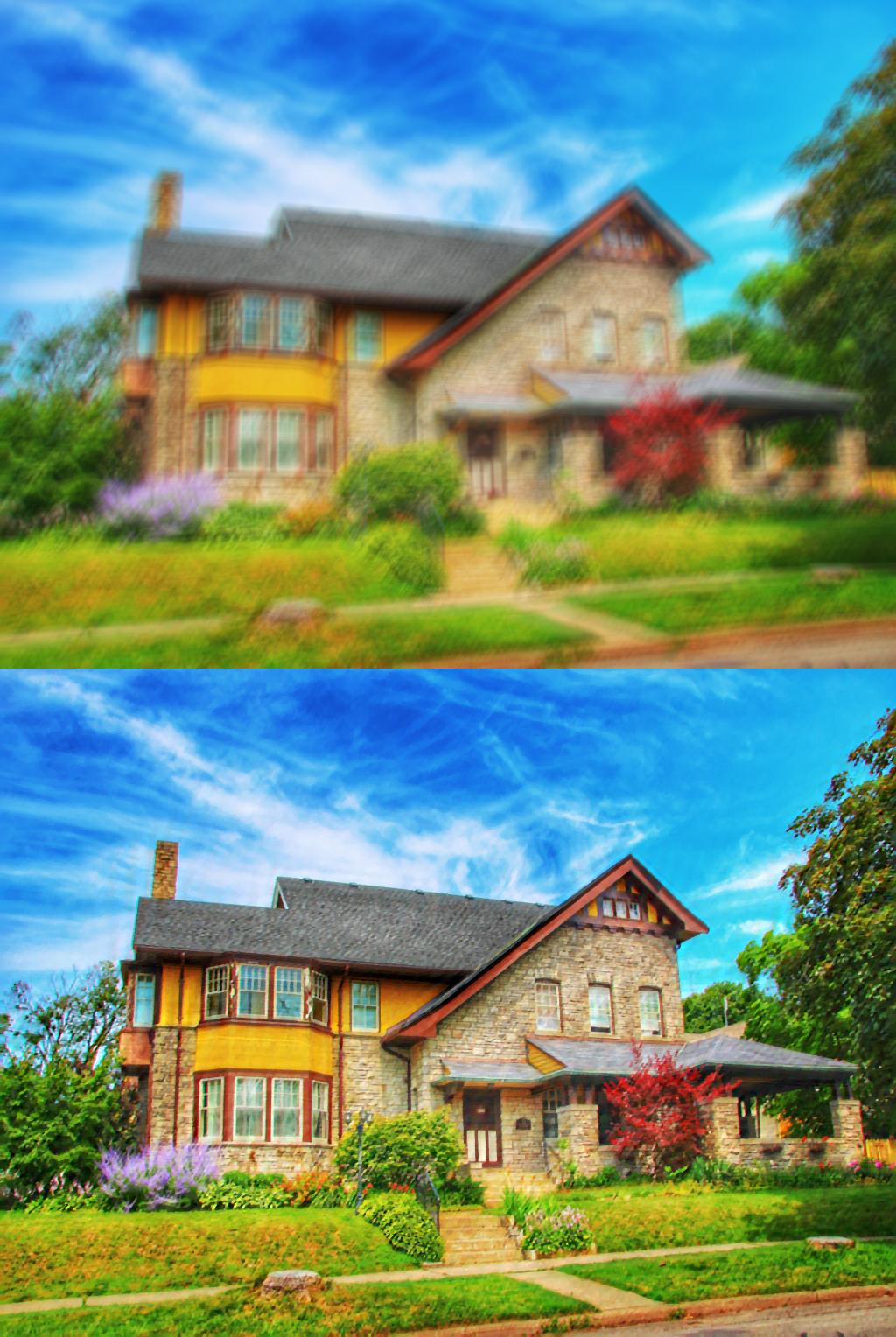}
	\includegraphics[height=1.3in]{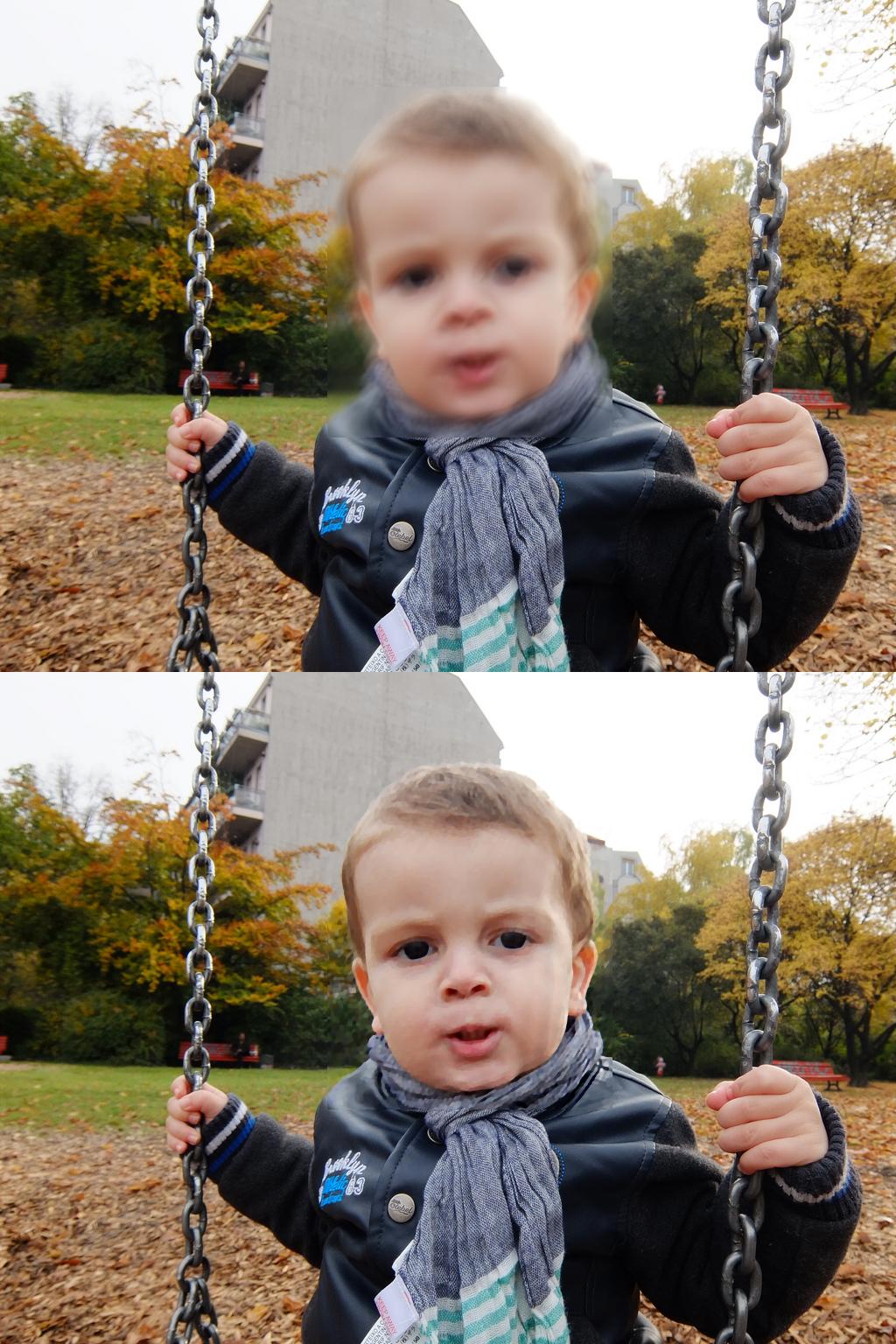}
	\includegraphics[height=1.3in]{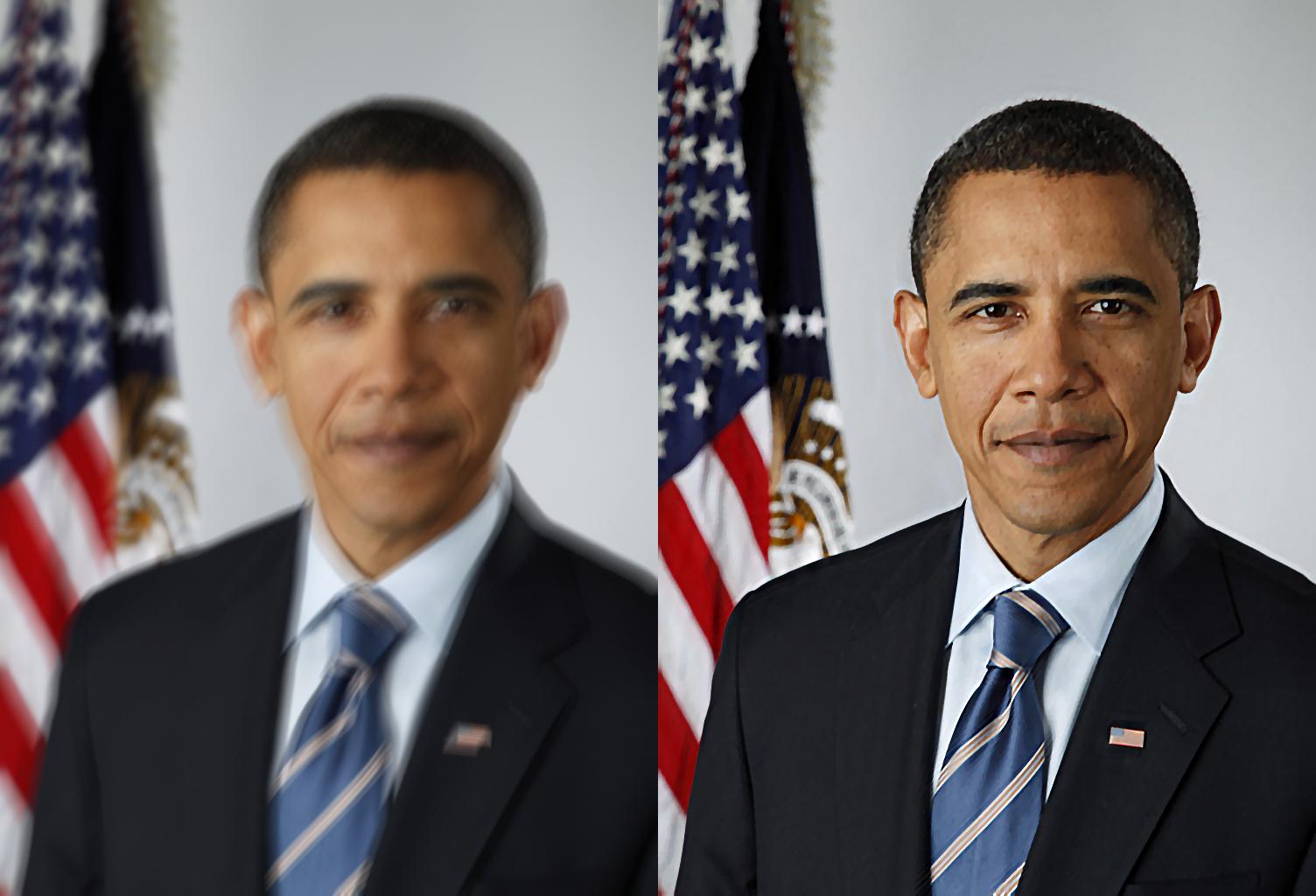}
\captionof{figure}{Given a blurred image (or blurred region), our algorithm PRIDA recovers the sharp image and the kernel associated with the blur.}
\label{fig:teaser}
\bigskip
}
\makeatother

\begin{document}

\title{Robust Blind Deconvolution via Mirror Descent}

\author{\centering Sathya N. Ravi \quad  Ronak Mehta \quad Vikas Singh\\
	{\tt\small \centering \hspace{25mm} \{sravi, ronakrm, vsingh\}@cs.wisc.edu} }

\maketitle


\input{abstract2.tex}
\input{intro.tex}
\input{prelim.tex}
\input{alg.tex}
\input{impl.tex}
\input{exps.tex}
\input{conclusions.tex}

\input{appendix.tex}
\bibliographystyle{natbib}
\bibliography{robust_blind}

\end{document}

%% file: abstract2.tex
\begin{abstract}
We revisit the Blind Deconvolution problem with a focus on 
understanding its robustness and convergence properties.
Provable robustness to noise and other perturbations is receiving recent interest in vision, from obtaining immunity to adversarial attacks to assessing and describing failure modes of algorithms in mission critical applications. 
Further, many blind deconvolution methods based on deep architectures 
internally make use of or optimize the basic formulation, so a clearer 
understanding of how this sub-module behaves, when it can be solved, 
and what noise injection it can tolerate is a first order requirement. 
We derive new insights into the theoretical underpinnings of blind deconvolution. 
The algorithm that emerges has nice convergence guarantees and is
provably robust in a sense we formalize in the paper. 
Interestingly, these technical results play out very well in practice, where 
on standard datasets our algorithm yields results competitive 
with or superior to the state of the art. 


\end{abstract}

%% file: intro.tex
\section{Introduction}
Image deblurring has been an active area of study in computer vision for nearly five decades. 
The early proposals 
sought to sharpen or {\em deblur} images from photographs
by relying on parameters relating the exposure and 
the amplifier gain, e.g., via the use of the Stroke/Zech division filter \cite{stroke1970new}. 
Most contemporary algorithms for deblurring, however, 
pose the problem as blind deconvolution, which refers to separating a true unknown signal and some unknown ``kernel" or ``filter" 
when provided knowledge {\em only} of the noisy measurement of the signal convolved with the filter.
This is a fundamental topic today in signal processing and vision, and remains challenging due to its
non-convex and ill-posed nature --- 
only within the last few years has brisk progress been made 
towards methods that gracefully handle real images encountered in practice \cite{levin2009understanding,campisi2016blind}.
These recent developments notwithstanding,
due to the foregoing technical challenges,
we are often unable to guarantee provably good solutions to the underlying optimization task, and 
strategies to address these issues are being studied by various researchers in our community today \cite{perrone2015blind,jin2017noise,li2016identifiability,campisi2016blind}. 

Modern approaches generally prefer one of two related but distinct strategies for blind deconvolution.
On the statistical side, research has primarily revolved around Bayesian methods \cite{ruiz2015variational}, taking advantage of useful priors ranging from fundamental image geometry in the context of
its relation to edge detection and saliency, to expert knowledge of the specific application domain of interest \cite{cho2009fast,xu2010two}. While these ideas
provide guarantees in terms of robustness, the development of efficient sampling (e.g., Gibbs sampler) and inference algorithms remains an active topic of research.
On the optimization side, total variation regularization has proven to be extremely effective in general image deblurring \cite{perrone2014total,chan1998total,osher2005iterative} in a variety of image domains.
While the mathematical properties of total variation have been well studied in applied mathematics, signal processing and machine learning, 
our understanding of the robustness and convergence behavior of even the best performing algorithms for blind deconvolution based
on this construct remains 
limited, although there is exciting progress being made \cite{Srinivasan_2017_CVPR}. A primary motivation of our work is to shed light on these theoretical issues.

Separate from, but complementary to the above lines of work, the enormous success of deep convolutional architectures in vision has led to a number of papers \cite{sun2015learning,schuler2016learning,chakrabarti2016neural,noroozi2017motion}
exploring how such successes can be adapted to deconvolution in general. While some initial attempts 
showed the use of machine learning methods for {\em non-blind} image deconvolution (i.e., the blur kernel is provided),
discriminatively trained architectures have now been shown to work quite well for the general setting, both with and without priors on motion blur types.
A natural question one may ask is whether an in-depth study of the core blind deconvolution formulation and its
properties is relevant in light of this still evolving body of convolutional neural networks based literature.
The reader will see that our work is complementary. Of the recent
proposals in this line of work reformulate deconvolution as a supervised learning problem by synthesizing blurred and sharp image pairs,
 and are often based on some form of blind deconvolution sub-routine internally \cite{schuler2016learning}. As these methods
get closer to practical deployment in mission critical applications, a detailed assessment of their behavior profile
will be a first order requirement for regulation compliance.
To enable investigating the robustness and convergence properties of these architectures and their resilience to adversarial examples --- as is happening
in the last few years for other problems in both computer vision and machine learning \cite{2017suonepixel,moosavi2016deepfool,moosavi2017universal} --- 
we will necessarily rely on and benefit from a ``first principles'' understanding of such properties for the standalone (i.e., shallow regimes of) blind deconvolution. 

{\bf Contribution.} In this paper, we provide (1) a {\em quantifiably and provably} {\bf robust} algorithm for blind deconvolution with (2) {\bf guaranteed convergence} properties.
To our knowledge, no algorithm is currently known that offers {\em both} these properties at once. Our convergence guarantees match the best
known results in optimization at this time. Our technical analysis is also backed up by practical performance. Via an extensive experimental study, we show that
on most available benchmarks, our simple algorithm competes favorably with (or is superior to) the state of the art, and provide a user-friendly implementation which can be easily extended to a complete user-interactive deblurring package.

\subsection{Prior Work}\label{sec:prior}
Methods for image deblurring via blind deconvolution have employed a variety of regularizations derived from a wide range of image priors. The literature is vast and so we
restrict our discussion to a subset of works that are closely related to or motivate our proposed strategy.
The earlier forms of regularization were based on the $\ell_2$-norm in \cite{you1996regularization}, where an alternating minimization scheme was proposed.
More recent improvements have been proposed by Cho and Lee \cite{cho2009fast} and Xu and Jia \cite{xu2010two}.
On the other hand, total variation regularization -- the defacto choice in many state of the art methods today -- was initially deployed in image denoising applications
\cite{rudin1992nonlinear,vogel1996iterative}.
and brought to the image deconvolution problem by Chan and Wong \cite{chan1998total}.
A nice result by \cite{osher2005iterative} gives a variational iterative procedure for solving the total variation objective. 
A conceptually distinct set of results for blind deconvolution adopt a more statistical approach instead. 
Levin et al. in \cite{levin2009understanding} provide analysis of algorithms following \textit{maximum a posteriori} (MAP) estimators.
A recent work \cite{ruiz2015variational} gives a nice and comprehensive overview of Bayesian methods for blind deconvolution. 
A few years back, \cite{perrone2014total} built on analysis in \cite{levin2009understanding} and demonstrated experimentally the behavior of \cite{chan1998total}
.
In a follow-up work, those authors showed the advantage of a logarithmic prior \cite{perrone2015blind}, obtaining state of the art results with a mild modification
to the classical TV-norm based formulation which we will present shortly.
Separate from total variation regularization based approaches, interesting results have been shown by \cite{michaeli2014blind} through an $\ell_0$ regularization
on text images and by \cite{michaeli2014blind,sun2013edge} via the use of patch priors. Recently, a detailed
comparative study was conducted by \cite{lai2016comparative}, in which participants were asked to qualitatively compare two results from multiple algorithms,
a subset of which are described in our review above.

In the last few years, ideas based on specialized deep networks have started yielding interesting results for this problem.
For example, \cite{sun2015learning} was among the first approaches for motion blur removal by posing the problem as a supervised learning task
and training a convolutional neural network (CNN) to infer the parameters. Schuler built on these results in \cite{schuler2016learning}, and Chakrabarti \cite{chakrabarti2016neural}
constructed a network to predict the Fourier coefficients of the filter necessary to deblur specific image patches.
Taking advantage of modern convolutional architectures, \cite{nah2016deep} constructed deep multi-scale networks for dynamic scene deblurring with strong empirical
results. In the past year, Generative Adversarial Networks have also been applied with measured success \cite{ramakrishnan2017deep}.

%% file: prelim.tex
\section{The Blind Deconvolution problem}\label{sec:prelim}
Throughout this paper we assume that an image is an $n-$dimensional vector taking values between $0$ and $1$ without loss of generality. We will use $f\in\R^n$ and $k\in\R^s$ to denote the vectorized sharp image and blur kernel, both of which are to be estimated given the vectorized blurry image $b\in\R^n$. Mathematically, the model can be written as,
\begin{align}
b = f* k + \alpha,
\end{align}
where $*$ denotes the usual convolution between two signals and $\alpha$ denotes the independent noise vector at each pixel. 
Assuming that $\alpha_i\sim\mathcal{N}(0,1)\ \forall i\in[n]$, we can estimate $f,k$ by maximizing the log-likelihood, corresponding to solving the following least squares optimization problem,
\begin{align}
\min_{f,k}\|f*k-b \|_2^2.\label{mle_blind}
\end{align}
Observe that the number of parameters to be estimated is $n+s$ and can be much larger than the number of observations $n$ if the kernel is large. To solve for solutions to \eqref{mle_blind}, many regularization functions (or priors) and/or constraints have been proposed in the literature \cite{levin2009understanding,ruiz2015variational,campisi2016blind}. To keep the presentation
simple, we will focus our attention on two generic components that have shown strong empirical performance to specify the full model.

{\bf Component 1)} The Total Variation (TV) $\ell_p$-norm on $f$ has been shown to promote smoothness of the estimated image \cite{chambolle1997image}.
The image TV norm is defined as some norm of its discrete gradient field over the image lattice $(\nabla_if(\cdot),\nabla_jf(\cdot))$:
\begin{align}
\|f\|_{TV}^{p}:=(\left(\|\nabla_if\|_p+\|\nabla_jf\|_p\right)^{p}.
\end{align}
Note that for $p \in \{1,2\}$, this corresponds to the classical anisotropic and isotropic TV norm respectively. Our theoretical analysis extends to any $p\geq 1$, but we will
assume that $p=2$ to describe our results. 

{\bf Component 2)} In order to define a reasonable constraint set, we appeal to the fundamentals of the image capture process.
Pixel values are explicitly a positive function of the photon count at a specific point on the image sensor, and so we
enforce the constraint that the kernel must be nonnegative.
Further, a blurred image can be interpreted as a weighted average of a sharp image captured with slight shifts, typically stemming from an extended exposure time due to
a variety of reasons.
Together, these requirements form our constraint set: the probability simplex $\Delta := \{x\geq0: 1^\top x = 1\}$.
With these two pieces, the problem that we aim to solve can be formally written as,
\begin{align}\label{our_blind_form}
\min_{f,k\in\Delta}&~\loss(f,k):=\|f*k-b \|_2^2 + \lambda\|f\|_{TV}
\end{align}
where $\lambda\geq 0$ is a tunable regularization parameter. Intuitively, higher values of $\lambda$ will encourage more smoothness in the optimal sharp
image $f$ of \eqref{our_blind_form}. 

%% file: alg.tex
\section{The PRIDA Algorithm}\label{sec:prida}
In principle, Problem \eqref{our_blind_form} should be easily amenable to many continuous optimization methods but
in practice, \cite{perrone2014total} provides compelling evidence that choosing the right algorithm is critical
to a successful recovery of the sharp image $f$. 
Notice two important but straightforward properties of the optimization in \eqref{our_blind_form}:
\begin{inparaenum}[\bfseries 1)]
\item the objective is smooth and convex in each argument $f,k$ individually but {\em not} jointly convex and
\item the feasible set $\Delta$ is convex and compact.
\end{inparaenum}

{\em Roadmap.} We will see shortly that properly exploiting these two simple properties will suggest a natural choice of an algorithm that is
familiar in non-linear optimization but not very broadly used in machine learning and vision. Interestingly, 
after we motivate the choice of the algorithm, we will see how the properties above provide certain technical results
that yield guarantees for fast convergence rates and subsequently, suggest strategies for a rigorous robustness analysis.
But first, let us analyze why some obvious simplifications and/or direct use of an alternating scheme may not be an effective
strategy for this model. 

{\bf Potential Idea: ignore nonconvexity?} A natural strategy to solve \eqref{our_blind_form} may be to use an algorithm
which exploits the convexity of  $\loss$ individually with respect to $f$ and $k$. A well known method that
offers this capability is the Alternating Minimization (AM) algorithm \cite{hardt2014understanding}. The AM algorithm for this model
performs the following calculation (or update) at each iteration:\begin{align}
&f^{t+1}\leftarrow \arg\min_f\loss\left(f,k^t\right)\label{am_f_update}\\
&k^{t+1}\leftarrow \arg\min_{k\in\Delta}\loss\left(f^{t+1},k\right)\label{am_k_update}.
\end{align}
Since both the subproblems \eqref{am_f_update} and \eqref{am_k_update} are convex optimization problems, they can both be computed efficiently \cite{hardt2014understanding}. 

{\em A potential problem of Alternating Minimization: Random versus structured blur.}
 There are some recent results that analyze the convergence behavior of the AM algorithm for \emph{random} blur kernels \cite{hardt2014understanding},
and offer guarantees on its performance. 
Unfortunately, it is still an open question
whether such guarantees are available for {\em structured blur kernels} that we universally encounter in vision.
In fact, \cite{perrone2014total} explicitly constructs an illustrative example where the AM algorithm converges to a strict saddle point due to the nonconvexity of $\loss$.

In the context of the blind deconvolution problem, strict saddle points correspond to a {\bf no blur solution},
that is, when the kernel $k$ has {\em only one nonzero entry}. We see that in \cite{perrone2014total} (cf. Section 3.4),
the authors give a clear example where the AM algorithm converges to the no blur solution, and thereby propose specific work-arounds to solve the subproblem \eqref{am_k_update}
such that the algorithm empirically converges to the desired one instead.
The authors also show that their scheme performs consistently better on many standard benchmark datasets. However, to our knowledge, it is not clear
if the procedure suggested in \cite{perrone2014total} guarantees convergence in general. Whether the method in \cite{perrone2014total} provably returns a minimizer of \eqref{our_blind_form}
is also not described in their work. 

{\bf Revisit Gradient Methods?}
Instead of the alternating scheme, we take a more ``classical'' approach to this problem and propose 
updating both $f$ and $k$ simultaneously at each iteration.
Our choice of algorithm, described shortly,
is motivated by {\bf two key insights} in Problem \eqref{our_blind_form}.
{\bf First}, for a smooth optimization problem, it has recently become known that the set of initial points from where
  a first order gradient method converges to a saddle point has a Lebesgue measure of zero \cite{panageas2016gradient}.
This immediately entails that with with very high probability, a gradient method will converge to a local minimizer.
{\bf Second}, the geometry of $\Delta$ will allow us to provably speed up the convergence which is interesting from both a theoretical standpoint and a practical one.
\begin{figure}[t]\label{fig:alg}
	\centering
	\includegraphics[width=3in]{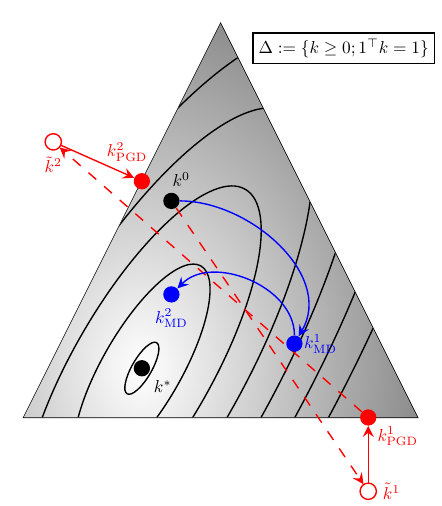}
	\caption{Visualization of Projected Gradient Descent (PGD) and Mirror Descent (MD) on the probability simplex. If the step size is not selected carefully, projected updates are more likely to find solutions along the  boundary.}
\end{figure}

{\bf A Mirror-descent style algorithm.}
To describe our algorithm, it is easiest to briefly review the form of a classical mirror descent (MD) scheme used in convex optimization. 
 Recall that the standard way to solve constrained optimization problems is to use projections, that is, we first take a (negative) gradient step and then a \emph{Euclidean} projection on to the feasible set, assuming that this is easy to do (as is the case with norm balls, hyperplanes and so on). This procedure is often referred to as
 Projected Gradient Descent (PGD):
 \begin{align}\label{eq:pgd}
 x^{t+1} = \Pi_{\Delta}(x^t - \eta_x g_x^t)
 \end{align}
 where $\Pi_{\Delta}$ is the Euclidean projection.
 Under mild conditions on the step size $\eta_x$, PGD in fact guarantees convergence.
 However, the use of PGD type algorithms require some caution: PGD completely disregards the geometry of the feasible set and only uses the local behavior of the objective function.  Hence, the algorithm can be very inefficient particularly in the high dimensional and large scale settings we see in vision \cite{mahadevan2012sparse,luong2012solving}. 

 Intuitively, Mirror Descent (MD) addresses this problem with the following simple modification: it is better to choose a function that acts like a metric \textit{depending on the feasible set}. This function is called the Distance Generating Function (DGF) and moreover, it is enough for that function to be a metric just on the feasible set \cite{juditsky2011first}. Exploiting this property,
 MD has been used to design algorithms that are provably faster than PGD \cite{nesterov2005smooth} and is the preferred algorithm in many applications \cite{srebro2011universality,jain2012mirror}. An excellent description of the MD algorithm and its variants is given in \cite{nemtut}. Recently, \cite{zhou2017stochastic} showed how to extend MD to a class of nonconvex problems called variationally coherent problems. But unfortunately, our problem  \eqref{our_blind_form} does not satisfy the assumptions, hence it is not clear how or if the results shown in \cite{zhou2017stochastic} apply.

 Motivated by the above discussion, we propose a Provably Robust Image Deconvolution Algorithm (PRIDA), shown in Alg. \ref{alg:prida}. As alluded to previously,
 PRIDA is similar in spirit to the MD algorithm in Convex Optimization. The main difference between the standard MD algorithm and PRIDA is that the step size is chosen independently for each coordinate. The intuition behind the step size rule can be seen as follows: if a coordinate $j$ of the filter (kernel) $k$ at the $t-$th iteration $k_i^t$ is large in magnitude, then we expect it to remain reasonably high at the $t+1-$th iteration. Our empirical results show that this is very effective in practice. Next, we show that PRIDA converges provably to a minimizer. 
\begin{algorithm}[!t]
	\caption{ \label{alg:prida} Provably Robust Image Deblurring Algorithm (PRIDA)}
	\begin{algorithmic}
		\STATE Pick a starting point  $k^0\in\Delta$, $f^0\in [0,1]^n$.
		\FOR{$t = 0,1,2,\cdots,T$}
		\STATE $(g_f^t,g_k^t)\leftarrow \left(\nabla \loss_f\left(f^t,k^t\right), \nabla \loss_k\left(f^t,k^t\right)\right)$
		\STATE $f^{t+1}\leftarrow f^t-\eta_fg_f^t $
		\STATE $\hat{k}^{t+1}_i\leftarrow k^t_i\exp\left(-\eta^t_{k_i}g^t_{k_i}\right)\quad $ \large(*)\normalsize
		\STATE $k^{t+1}\leftarrow \frac{\hat{k}^{t+1}}{\sum_{i=1}^s\hat{k}^{t+1}}$
		\ENDFOR
	\end{algorithmic}
\end{algorithm}
\section{Convergence, Robustness, and More}\label{sec:conv}
To analyze PRIDA, we use the following equivalent interpretation of the update step (derived in the supplement):\begin{align}
f^{t+1} &= \arg\min_f  \langle g^t_f,f\rangle + \frac{1}{2\eta_f}\| f-f^t\|_2^2 \label{f_update} \\
\hat{k}^{t+1} &= \arg\min_{\hat{k} } \langle \eta^t_k \circ g^t_k,\hat{k} \rangle + \KL \left(\hat{k}||k^t\right)\label{k1_update}\\
k^{t+1}&=\arg\min_{k\in\Delta}\KL\left(k||\hat{k}^{t+1}\right)\label{k2_update}
\end{align}
where $\KL(x||y)$ represents the usual Kullback-Leibler divergence between $x$ and $y$, $\langle,\cdot,\cdot\rangle$ is the inner product, and $\circ$ denotes element-wise multiplication. Note that when the $KL$ divergence function is replaced by the Euclidean norm, the algorithm becomes the standard PGD update. Observe that $KL(x||y)$ acts as a distance-generating function on simplex $\Delta$, and hence $k^{t+1}$ is unique. In order to show convergence we use the following intermediate result.
\begin{lemma}\label{triangle_lemma}
	Let $x^*=\arg\min_{x\in\Delta} \langle z,x\rangle+\KL(x||x^0)$. Then for any $y\in \Delta,z\in\R^s$, we have that, \begin{align}
	\langle z,y\rangle+ \KL(y||x^0)\geq \langle z,x^*\rangle +\KL(x^*||x^0) + \KL(y||x^*).
	\end{align}
	\begin{proof} See supplement.
	\end{proof}
\end{lemma}
With this in hand, we have the following convergence result.
\begin{theorem}\label{prida_conv}
	Let  $\|\nabla^2\loss\|\leq L$, then with step sizes $\eta^t_{f},\eta^t_{k_i}\leq \min\left(\frac{\alpha \|k\|_{\infty}}{k_i^t\|g_k\|_{\infty}},\frac{1}{L}\right)$ where $1>\alpha>0$ is fixed, PRIDA converges to a local minimizer of $\loss$ (avoids strict saddle points) almost surely. 
\begin{proof} We will assume without loss of generality for the analysis that the step size $\eta=\eta^t_{f}=\eta^t_{k_i}=1/L$. We prove the convergence in two steps. In step 1, we show that the iterates of the PRIDA algorithm \ref{alg:prida} converges to a fixed point. In step 2, we show that there is a subsequence that converges to a stationary point, that is, a point that approximately satisfies the first order necessary conditions. We then use \cite{2017arXiv171007406L} to show that such a stationary point is a locally optimal solution. 

Step 1: For notational convenience, let $p:=[f,k]$, where the first $n$ coordinates denote $f$ and the last $s$ coordinates denote $k$ respectively. Define $h(p):= \loss(p) +I_{\Delta}(p)$ where $I_{\Delta}(p)$ is the indicator function that takes the value $0$ if $p_k:=p_{[n+1:n+s]}\in\Delta$ and $\infty$ otherwise. Then for any $x,y \in \R^{n+s}$, we have that,\begin{align}
h(x)&=\loss(x)+I_{\Delta}(x)\leq \loss(y) + \nabla\loss(y)^T(x-y) + \frac{L}{2}\|x-y\|_2^2+I_{\Delta}(x)&\label{conv_eq_1}\\
&  = \loss(y) + \nabla\loss(y)^T(x-y) + \frac{L}{2}\|x_f-y_f\|^2_2 + \frac{L}{2}\|x_k-y_k\|_2^2+I_{\Delta}(x)&\label{conv_eq_2}\\
&\leq \loss(y)+ \nabla\loss(y)^T(x-y) + \frac{L}{2}\|x_f-y_f\|^2_2 + \frac{L}{2}\|x_k-y_k\|_1^2+I_{\Delta}(x)&\label{conv_eq_3}\\
&\leq \loss(y)+ \nabla\loss(y)^T(x-y) + \frac{L}{2}\|x_f-y_f\|^2_2 + \frac{L}{4}KL(x_k||y_k)+I_{\Delta}(x)=:u(x,y)&\label{conv_eq_4}
\end{align}
where \eqref{conv_eq_1} is by smoothness of the gradient (assumption), \eqref{conv_eq_3} is by Cauchy-Schwarz inequality and \eqref{conv_eq_4} is by Pinsker's inequality (see page 88 in \cite{Tsybakov:2008:INE:1522486}). Note that the minimizer of $u(x,y)$ with respect to $x$ exactly corresponds to the update rule in PRIDA and that $u(x,y)$ is strongly convex in $x$ (again due to Pinsker's inequality, see page 301 in \cite{bubeck2015convex}). Hence we can bound the per iteration improvement by,\begin{align}
h(p^t)-h(p^{t+1}) &\geq h(p^t) - u(p^{t+1},x^t) = u(p^t,p^t)-u(p^{t+1},p^t)\\
&\geq \frac{L}{2}\|p_f^{t+1}-p_f^t\|^2_2 + \frac{L}{4}KL(p_k^{t+1}||p_k^t)\geq \frac{L}{2}\|p^{t+1}-p^{t} \|_2^2\label{conv_eq_5}
\end{align}
where inequality \eqref{conv_eq_5} follows from \eqref{conv_eq_4} . Summing up the inequalities (over $t$) in \eqref{conv_eq_5}, we see that the $\lim_{t\to \infty} \|p^{t+1}-p^{t}\|=0$, that is, algorithm converges to a fixed point.

Step 2: Since the update rule for $f$ is standard gradient descent, we know that the iterates converge to a point where the gradient vanishes,  see section 1.2.3. in \cite{nesterov2013introductory}. So we focus on the update rule for the $k$ for which  we use Lemma \ref{triangle_lemma}. Taking $z=\eta \left(\loss\left(p^t\right) + g_k^{t^T}\left(p_k-p_k^t\right)\right)$ there, we have that,\begin{align}
\KL (p_k^*||p_k^{t+1})&\leq \KL (p_k^*||p_k^{t}) + \eta g_k^{t^T}(p_k^*-p_k^{t+1}) - \KL (p_k^{t+1}||p_k^t)\\
&=\KL (p_k^*||p_k^{t})+ \eta g_k^{t^T}(p_k^*-p_k^{t}) + \eta g_k^{t^T}(p_k^t-p^{t+1}) - \KL (p_k^{t+1}||p_k^t)\nonumber\\
&\leq \KL (p_k^*||p_k^{t})+ \eta g_k^{t^T}(p_k^*-p_k^{t}) +\loss(p^t)-\loss(p^{t+1})\label{conv_eq_6}
\end{align}
where we used the smoothness assumption in \eqref{conv_eq_6}. Again summing the inequalities in \eqref{conv_eq_6} (over $t$), we have that,\begin{align}
\KL(p_k^*||p_k^{t+1})\leq& \KL(p_k^*||p_k^{0}) +\eta \sum_{t} g_k^{t^T}(p_k^*-p_k^t)+\loss (p^0)\\
\implies T \min_t g_k^{t^T}(p_k^t-p_k^*)\leq & \sum_{t} g_k^{t^T}(p_k^t-p_k^*)\leq (\KL(p_k^*||p_k^{0})+\loss (p^0))L.
\end{align}
 Taking the limit as $T\to\infty$, we showed that we can find a point that satisfies the first order optimality conditions of our optimization problem. Thus we have shown that after $T$ steps, we can find a point that is $O(1/T)$ optimal.  PRIDA iterates now satisfy the assumptions of Proposition 10 in \cite{2017arXiv171007406L}, and so by Corollary 7 therein it directly follows that PRIDA does not converge to a strict saddle point almost surely.
\end{proof}
\end{theorem}
While we can get the same convergence rate (up to logarithmic factors in $s$) of $O(1/T)$  as that of PGD (see \cite{ghadimi2016accelerated}), the efficiency of PRIDA comes from the fact that   each iteration of PRIDA takes $O(s)$ time, compared to the $O(s\log s)$ required in PGD (see \cite{chen2011projection} for details) and is trivially parallelizable/amenable to GPU implementation. Details are included in the supplement. 
\subsection{Robustness}\label{sec:robust} Having shown the convergence of PRIDA, the natural follow-up investigation is to characterize its behavior in terms of its noise tolerance. 
We call an algorithm robust if it produces the same output on two different images such that one of them is a slightly perturbed version of the other. This notion of robustness has been recently introduced in the machine learning literature under the context of algorithmic stability \cite{hardt2015train}. Recent results in our community show that this is a critically desirable property of algorithms used in vision-based deployments since they are often sensitive to very small perturbations \cite{moosavi2017universal,2017suonepixel}. 

\textit{Plan of Attack.} Using only the main concepts of stability, we aim to measure the robustness of our algorithm. In typical stability analyses, noise is often introduced in the gradient computation, as a proxy for stochastic or approximate gradient updates. We follow this idea, and aim to bound the difference between the result of a noisy gradient update and a clean one. To be specific, we assume that two images, one with noise and the other without, produce gradients that are approximately the same.

Hence, at iteration $t$ we observe some noisy gradient $\tilde{g}_k^t$ of $g_k^t$ (and respectively $\tilde{g}_f^t$ of $g_f^t$). We would like to bound the distance between $k^{t+1}$ and $\tilde{k}^{t+1}$ ($f^{t+1}$ of $\tilde{f}^{t+1}$). In what follows, we look only at the update for the kernel $k$, but note that an analogous argument can be made for the sharp image $f$: the update step for $f$ is essentially a (sub)gradient step, and so the argument is simpler.
	\begin{lemma}\label{robust_analysis}
	Let $k^0$ be the initial point where all coordinates are equal. Let $g:=g_k^1$ be the true gradient and $h:=g^1_k+\alpha$ be some noisy gradient. Then, we have that $k^1_g,k^1_h$ computed using $g$ and $h$ are $\delta$-close in the $\ell_1$ sense.
	\begin{proof}		
		In order to study the robustness properties of our algorithm, we will use the interpretation of PRIDA given in \eqref{k1_update} and \eqref{k2_update}. Because the noisy gradient is only being used in the \eqref{k1_update}, we analyze how much iterates can stray after each of the two updates separately. To that end define the intermediate iterate $x:=\hat{k}^1_g$ computed using the true gradient and similarly $y:=\hat{k}^1_h$ the noisy one. To make the proof simple, we will assume that the step size is same for all the coordinates, that is, $\eta^1_i\equiv\eta$ (say $1/L$) and note that the argument can be easily extended for the general case. Then, the distance between $x$ and $y$ can be bounded as follows,\begin{align}
				\|x-y\|_1 =\sum_i^n \left| k^0_i e^{-\eta g_i} - k^0_i e^{-\eta (g + \alpha)_i}\right| &=\sum_i^n k_i^0 e^{-\eta g_i} \left|1 -  e^{-\eta\alpha_i}\right|\\
				& \leq \sum_i^n k_i^0 e^{-\eta g_i} |\eta\alpha_i| \leq \eta \bar{\alpha} \|x\|_1 \label{rob_first_step}
			\end{align}
			where the first step follows from the definition of $x$ and $y$, and the last two from the fact that $e^{-x}\geq 1-x~\forall x$ and  $\bar{\alpha}:=\max_i\left|\alpha_i\right|$. Now we show that the second step of the update, which corresponds to a simple normalization, is also well behaved:\begin{align}
			\left\|\frac{x}{\|x\|_1}-\frac{y}{\|y\|_1}\right\|_1 &= \left\|\frac{x}{\|x\|_1}-\frac{y}{\|x\|_1}+\frac{y}{\|x\|_1}-\frac{y}{\|y\|_1}\right\|_1\\
			&\leq \left\|\frac{x}{\|x\|_1}-\frac{y}{\|x\|_1}\right\|_1 + \left\|\frac{y}{\|x\|_1}-\frac{y}{\|y\|_1}\right\|_1\label{rob_tri}\\
			&=\frac{\|x-y\|_1}{\|x\|_1} + \left\|\frac{y\left(\|y\|_1-\|x\|_1\right)}{\|x\|_1\|y\|_1} \right\|_1\\
			&=\frac{\|x-y\|_1}{\|x\|_1} + \frac{\left|\|y\|_1-\|x\|_1 \right|}{\|x\|_1}\leq \frac{2\|x-y\|_1}{\|x\|_1}\label{rob_rev_tri}\\
			&\leq 2\eta \bar{\alpha}\label{rob_from_first_step}
			\end{align}
			where we use the triangle inequality for \eqref{rob_tri}, the reverse triangle inequality for the inequality in \eqref{rob_rev_tri}, and \eqref{rob_from_first_step} follows from \eqref{rob_first_step}. 
			If the noise level satisfies  $\bar{\alpha}\leq \delta/2\eta$, then we know the iterates computed using the noisy and true gradients are at most $\delta$ away.	
\end{proof}
	\end{lemma}
	\begin{remark} This result clearly shows the interplay between the noise level $\alpha$ and the step size $\eta$. When a sharp image $f$ undergoes convolution followed by the addition of noise, Lemma \ref{robust_analysis} tells us that it is better to take short steps instead of being overtly aggressive. 
\end{remark}
\begin{remark} \textit{Why are short steps sufficient in practice?}
  Given that every pixel in the blurred image is a nonnegative combination of neighboring pixels in the sharp image, it is enough to search among its \textit{neighbors} to form a realistic image rather than searching over the whole image space. This can be performed efficiently using short steps.
	\end{remark}

%% file: impl.tex
\subsection{Implementation Details}\label{sec:impl}

\textbf{Initialization.}
We follow the standard practice common across many vision problems and estimate both $f$ and $k$ at many resolutions. More specifically, our estimation proceeds through a coarse-to-fine pyramid scheme. For each level, we run PRIDA (Algorithm \ref{alg:prida}) and upscale the resulting estimated image and kernel for the next level.

 At the coarsest level, we initialize the kernel to be uniform, that is, $k_i^0=1/s$. While this choice of initialization is critical for many existing algorithms  \cite{perrone2014total,pan2014deblurring}, it is not  so important for PRIDA. Because the objective function is (jointly) bilinear, it may be the case that the initial few gradient steps will push some of the coordinates of the kernel to $0$ after a Euclidean projection. This is problematic because it will remain at $0$ during the entire course of optimization (at that scale), thus reducing the pyramid scheme's effectiveness. 
PRIDA on the other hand can be thought of as a version of ``soft-removal'': the multiplicative nature will naturally force all elements of any given kernel to remain strictly positive at all times, and hence a few ``bad'' steps will not necessarily hurt the overall performance.

\begin{figure}{!b}
	\centering
	\includegraphics[height=2.5in]{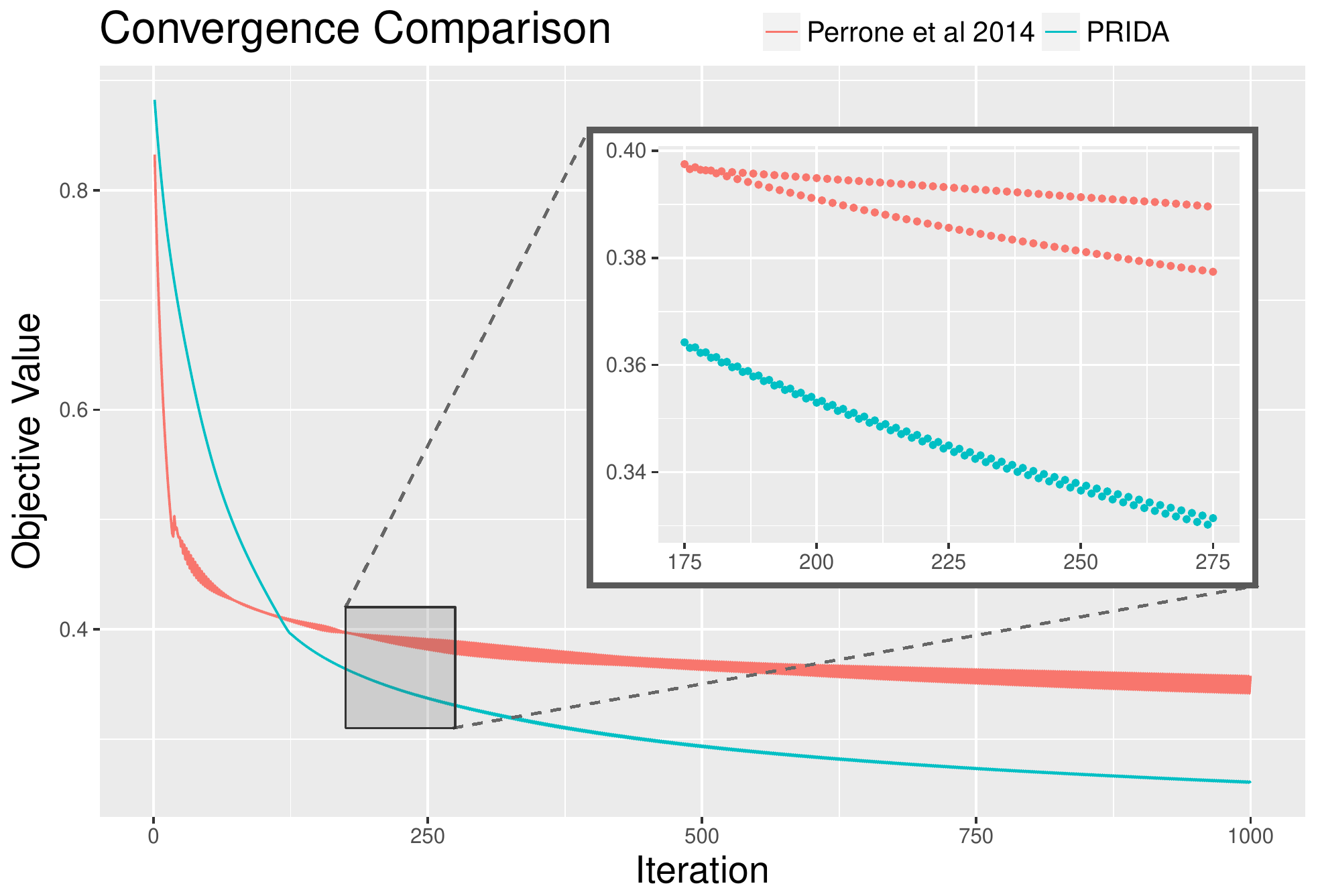}
	\caption{\label{fig:convplot}Convergence rate comparison of PRIDA (in red) and Perrone et al. in \cite{perrone2014total} (in blue).}
\end{figure}

\textbf{Numerical Considerations.}
When calculating the step size per pixel $\eta_{k_i}$, it may be the case that a given point in the kernel has already been driven close to 0. In this case if the (noisy) gradient is negative, however small, the computed step $e^{-\eta_{k_i} g_{k_i}}$ may be $+\infty$ if the value at that point has fallen below machine precision. To avoid these issues, we apply a ``Big $M$" correction
\cite{nemirovski2004interior} (chapter 4) such that the step taken is the minimum of $\{\exp\{-\eta_i g_i\},M\}$, where $M$ is a large positive constant. Intuitively, a large $M$ will allow PRIDA to take larger steps, thus encouraging faster convergence.  We fix $M=1000$ throughout our experiments.

%% file: exps.tex
\section{Experimental Evaluation}\label{sec:exps}

\begin{figure*}[!t]	
	\centering	
	\begin{tabular}{*5c}		
		\multicolumn{5}{c}{\includegraphics[width=0.95\textwidth]{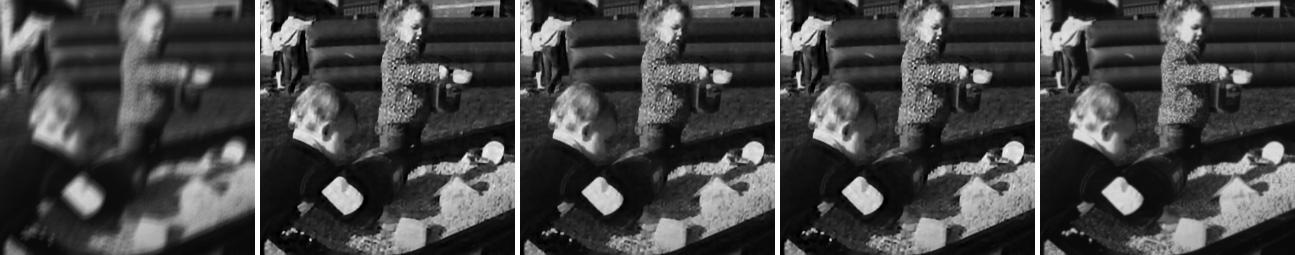}}\\
	\multicolumn{5}{c}	{\includegraphics[width=0.95\textwidth]{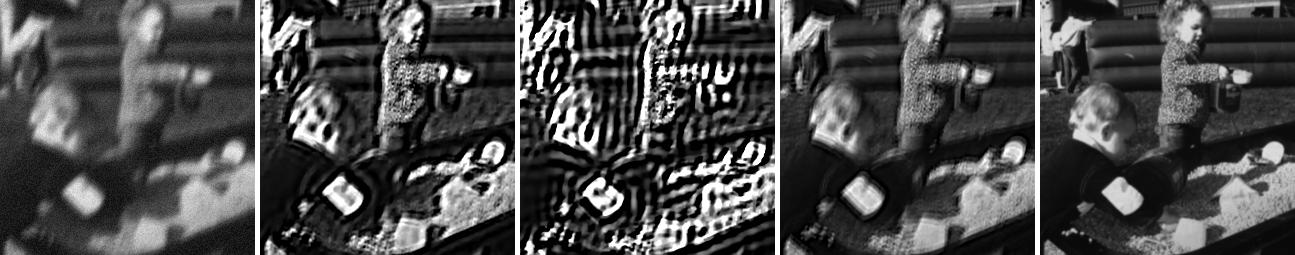}}\\
	\multicolumn{5}{c}	{\includegraphics[width=0.95\textwidth]{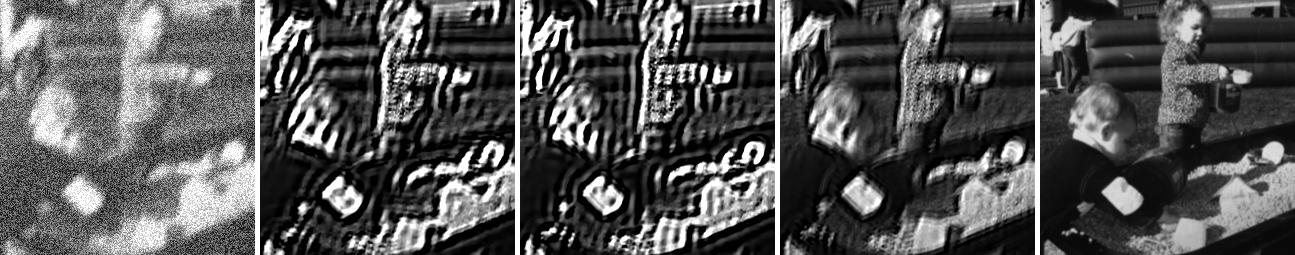}}\\
		\label{fig:levin1}
	\end{tabular}
	\caption{From left to right: (a) Input blurred images with added Gaussian noise. (b) Result from \cite{perrone2014total}. (c) Result from  \cite{perrone2015blind}. (d) Our Result. (e) Ground Truth. From top to bottom, each row corresponds to added noise with standard deviation 0,0.1, and 0.5 respectively.}\label{fig:levin}
	\end{figure*}

All experiments were conducted using MATLAB 2017a running on a 12-core Xeon E5-2620 @ 2.4 GHz machine with 64GB RAM. For all experiments on images of size $255 \times 255$, we use a fixed regularization hyperparameter of $\lambda=6e^{-4}$. The run time of each image on the finest scale is approximately $2$-$3$ minutes. In the first two sets of experiments, our goal is to validate the theoretical properties of PRIDA shown in earlier sections viz., convergence and robustness. Finally, we test if PRIDA is efficient on real world color images.  We compare with two recent standard baselines that are closely related to our algorithm \cite{perrone2014total,perrone2015blind}, and provide additional experimental details and comparisons with other algorithms in the supplement.

\subsection{Convergence}
Figure \ref{fig:convplot} shows the function value convergence rates for PRIDA and for \cite{perrone2014total}. Using the same pyramid scheme, we compute the function value for 1000 iterations of both algorithms over the finest level, fixing $\lambda$ as stated above for PRIDA and the default setting provided by the authors in \cite{perrone2014total}. Notice that while \cite{perrone2014total}'s method initially drops quickly, our method eventually converges much faster to a lower objective function. We note also that the PRIDA updates are significantly more stable, providing evidence of our robustness analysis above.
\begin{figure*}[!t]
	\centering
	\includegraphics[width=0.19\textwidth]{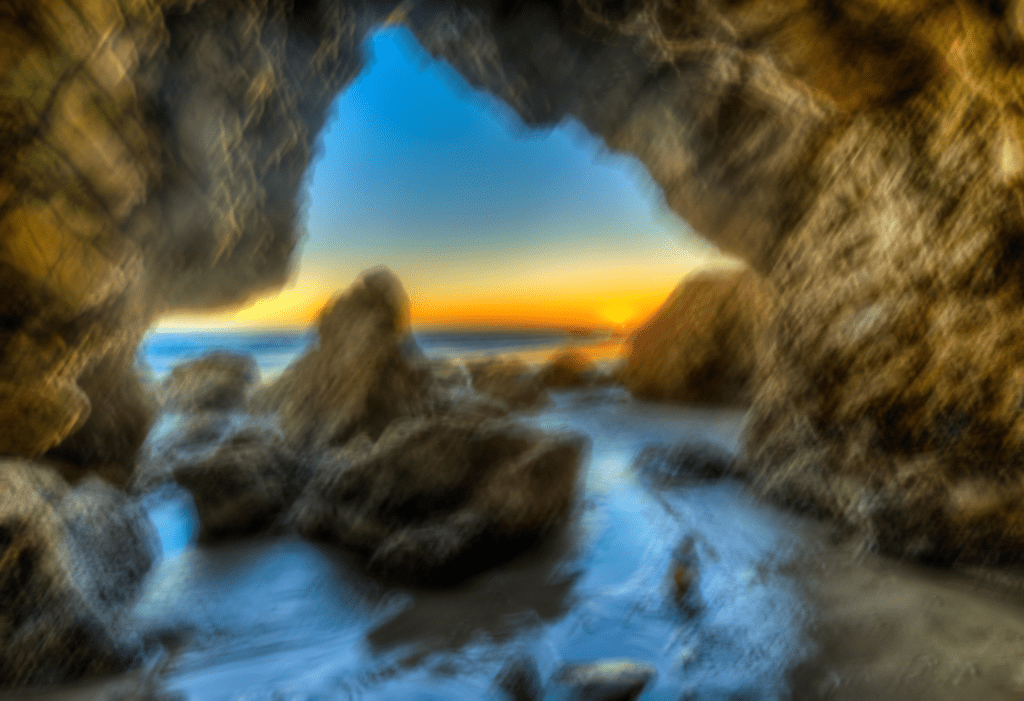}
	\includegraphics[width=0.19\textwidth]{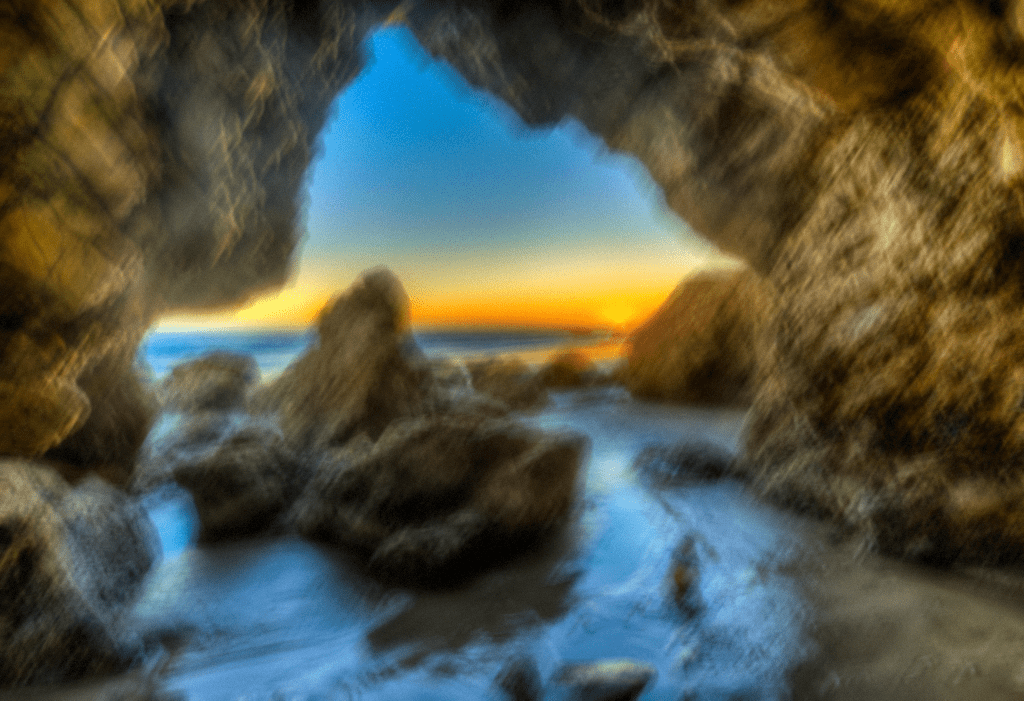}
	\includegraphics[width=0.19\textwidth]{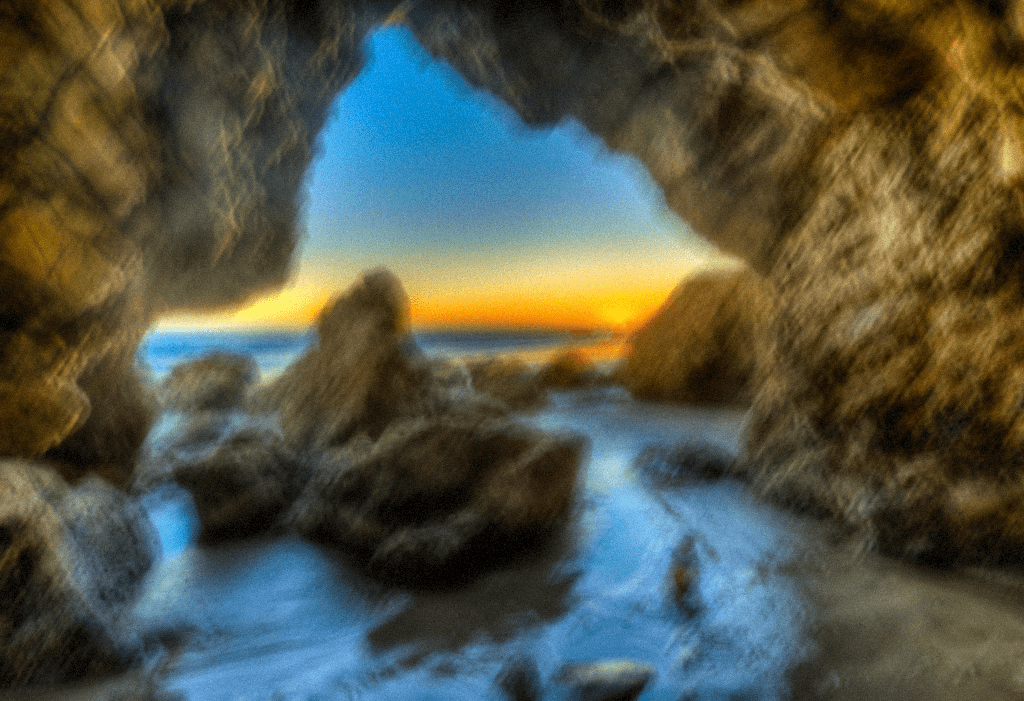}
	\includegraphics[width=0.19\textwidth]{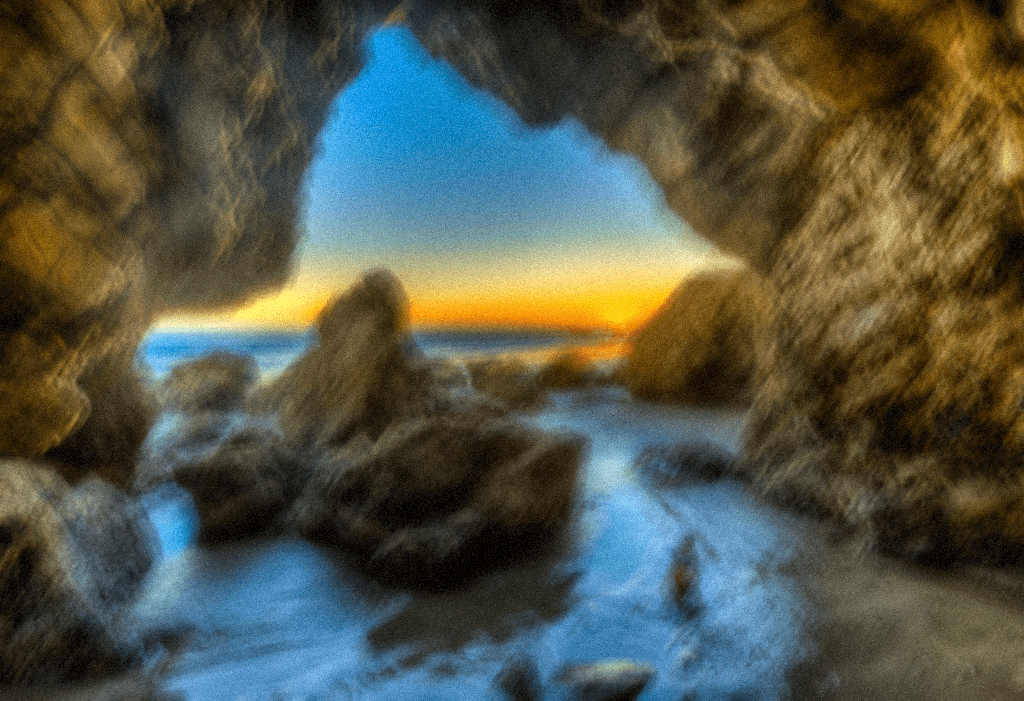}
	\includegraphics[width=0.19\textwidth]{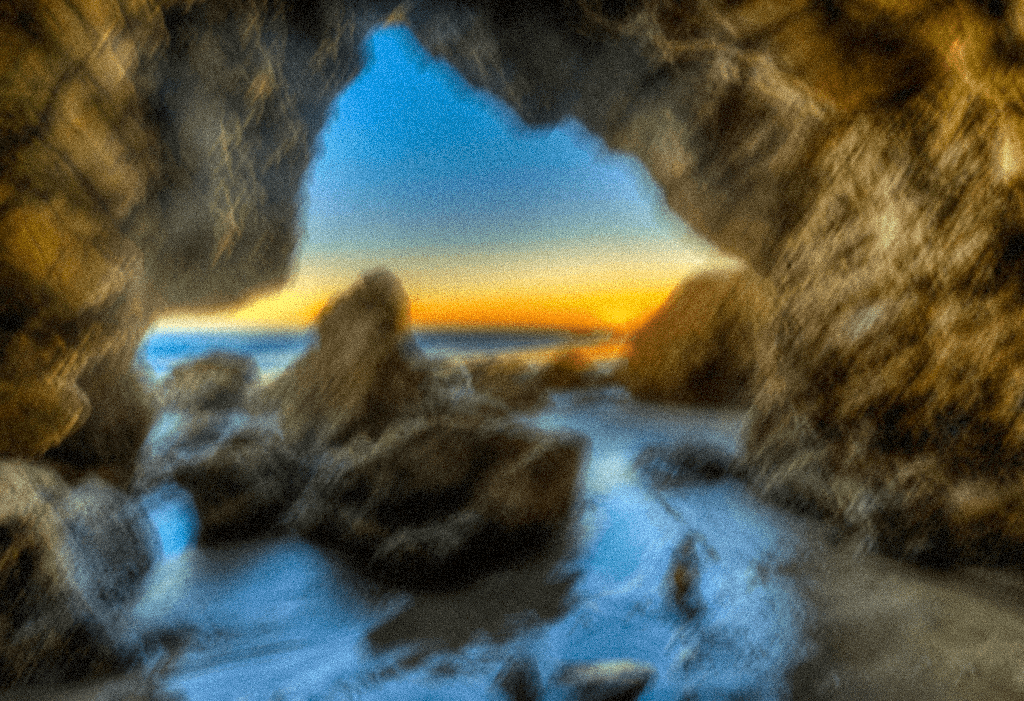}
	
	\includegraphics[width=0.19\textwidth]{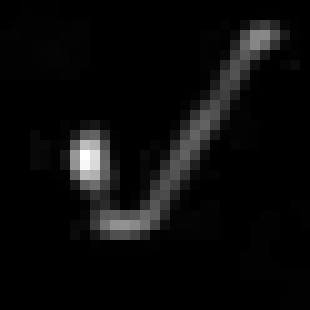}
	\includegraphics[width=0.19\textwidth]{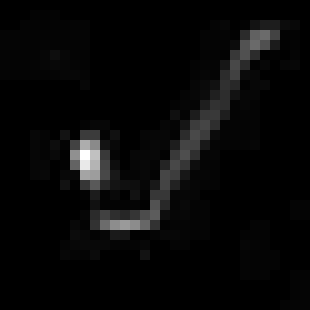}
	\includegraphics[width=0.19\textwidth]{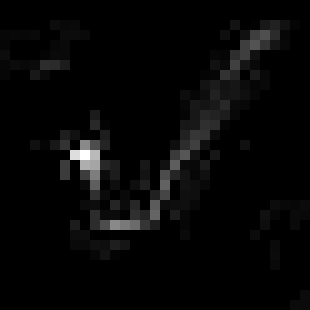}
	\includegraphics[width=0.19\textwidth]{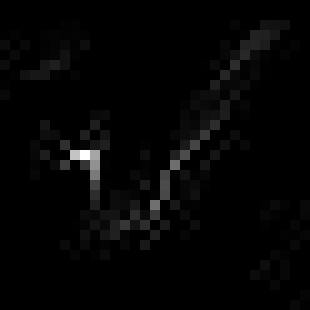}
	\includegraphics[width=0.19\textwidth]{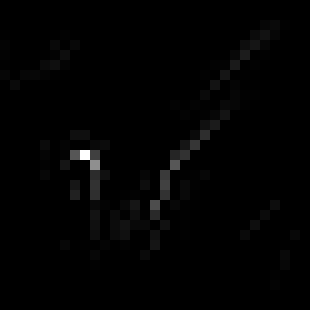}
	
	\includegraphics[width=0.19\textwidth]{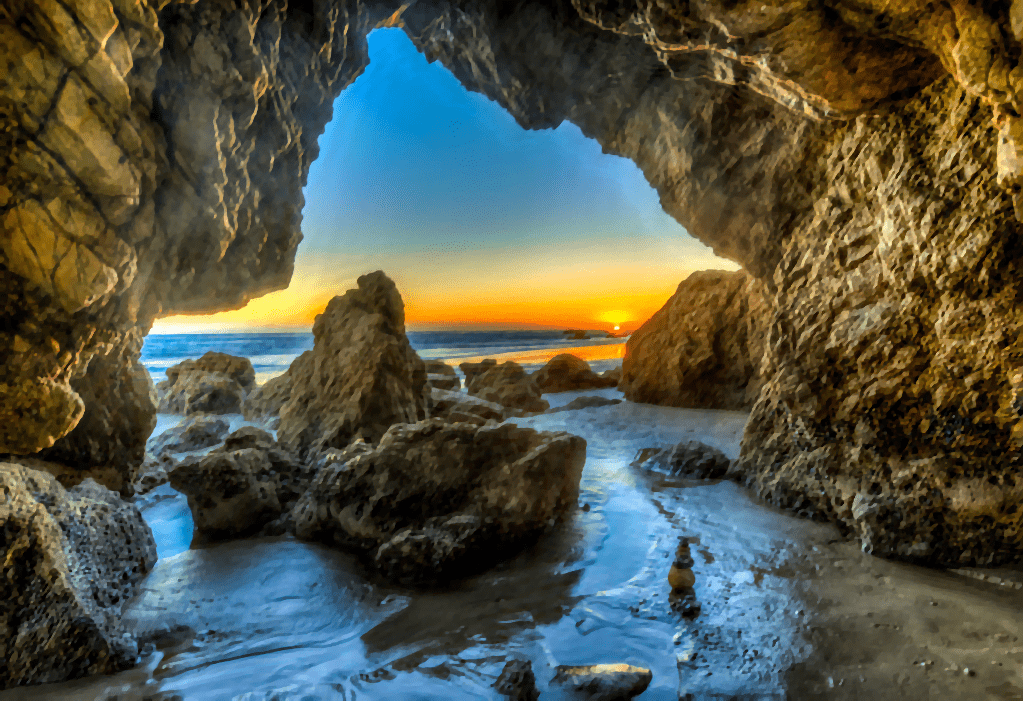}
	\includegraphics[width=0.19\textwidth]{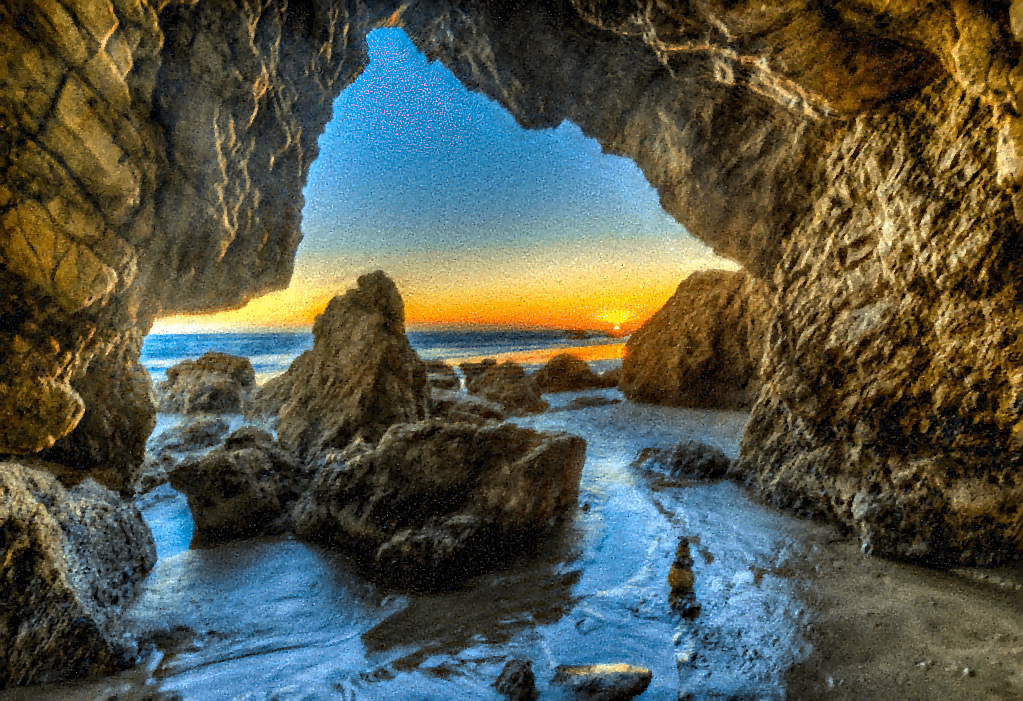}
	\includegraphics[width=0.19\textwidth]{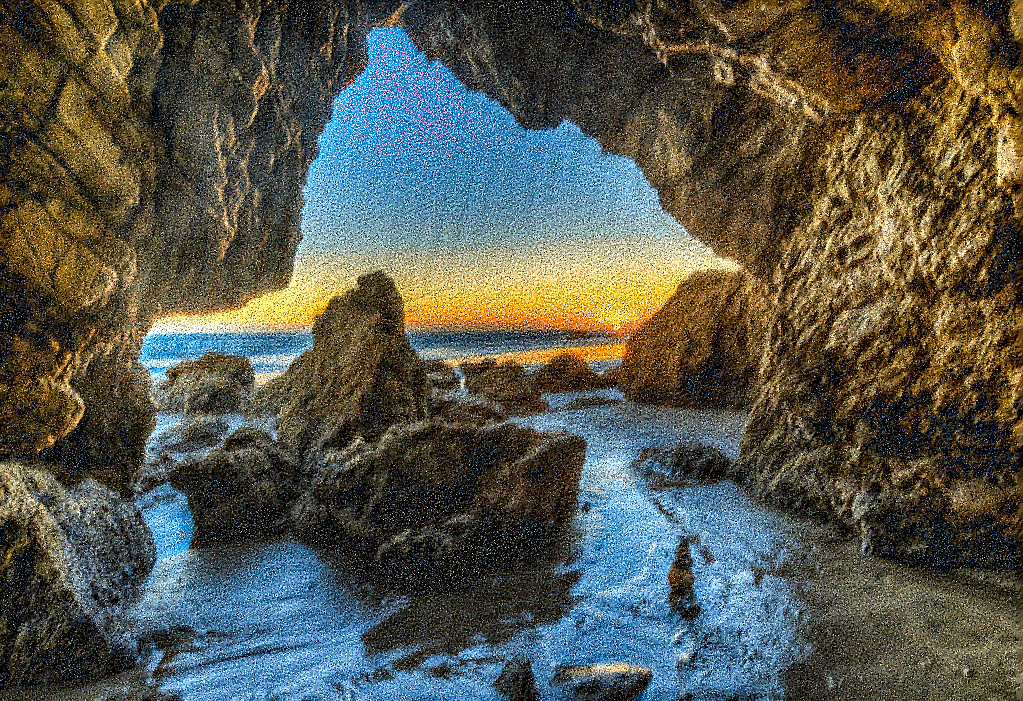}
	\includegraphics[width=0.19\textwidth]{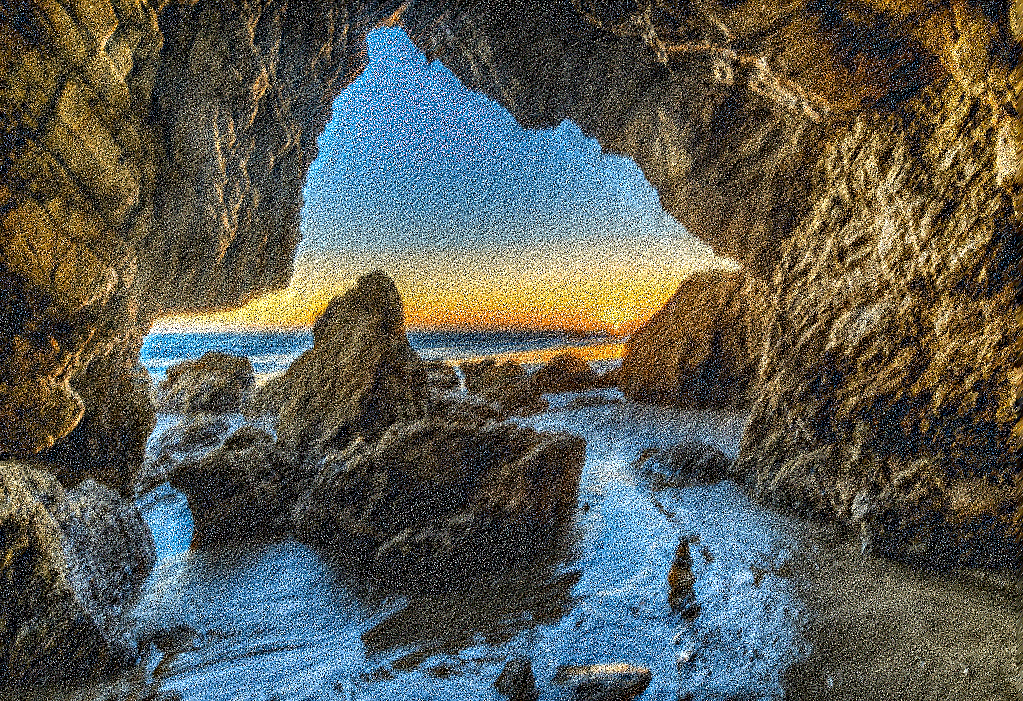}
	\includegraphics[width=0.19\textwidth]{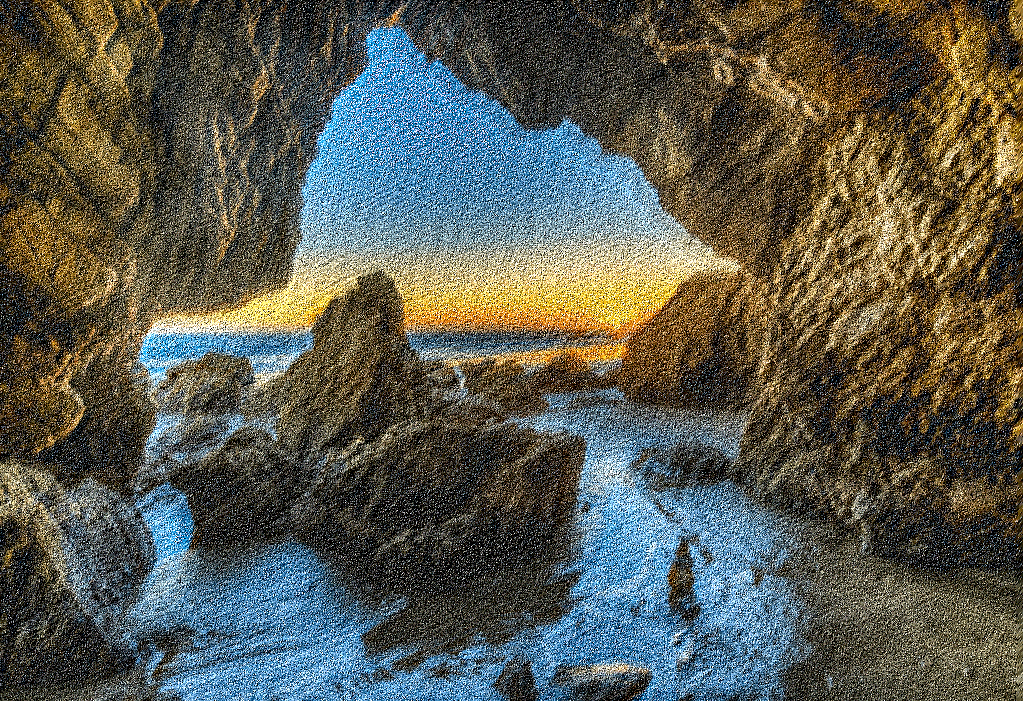}
	
	\caption{Color image recovery in the presence of intensity noise. From left to right, we add 0-mean Gaussian random noise with variance $0,1,4,9,16$ respectively. The first row shows the blurred and noisy input, the second the recovered kernel, and the third our final image recovery. Standard denoising methods can be applied to the deblurred image. }\label{fig:colornoise}
\end{figure*}
\subsection{Robustness}
\begin{table}[!t]
	\centering	
	\begin{tabular}{cccccc}
		\toprule\toprule
		& \multicolumn{5}{c}{Noise Level $\sigma$}   \\ 
		Alg.  & 1      & 3      & 5      & 7      & 9      \\ \hline
		\cite{perrone2014total} & 0.0008 & 0.0223 & 0.0584 & 0.0849 & 0.0957 \\
		\cite{perrone2015blind} & 0.0006 & 0.0994 & 0.1375 & 0.1212 & 0.0941 \\
		PRIDA & 0.0008 & 0.0041 & 0.0089 & 0.0149 & 0.0231\\
		\bottomrule\bottomrule
	\end{tabular}
	\smallskip
	\caption{Average endpoint error across the dataset in \cite{levin2009understanding} for varying levels of noise.}\label{fig:table}
\end{table}
To exemplify the robustness of PRIDA to noise, we conduct experiments on the well-known dataset first introduced by  \cite{levin2009understanding}. The grayscale images are $255 \times 255$ pixels in size with known blur kernels ranging in size from 13 to 27 pixels square. To evaluate robustness, we add varying levels of noise to each image, and qualitatively evaluate the end result. We compare our method to the algorithms presented in \cite{perrone2014total} and in \cite{perrone2015blind}.

We show the results of PRIDA in comparison with the standard baselines in Figure \ref{fig:levin}, see supplement for more results. To generate the noisy and blurred images, Gaussian random noise with mean $\mu=0$ was added to each blurred image. Here we can clearly observe the ability of our procedure to handle large amounts of noise. Over the entire dataset, we observe that in some interesting cases both algorithms from \cite{perrone2014total} and \cite{perrone2015blind} are able to recover a reasonably sharp image in the presence of noise. Over the entire dataset from \cite{levin2009understanding}, however, we note that their results are significantly more variable than that of PRIDA. On average, PRIDA is much more consistent in recovery over the entire dataset, shown in Table \ref{fig:table}, validating our theoretical analysis above.

\subsection{Scaling Up to the Real World}
While the results above are valuable in validating our theoretical claims, we also evaluate our algorithms' robustness on real world images.  Computationally, an interesting property of PRIDA is that all of its operations involve convolutions (Fast Fourier Transforms) and elementwise operations, both of which can benefit from GPU efficiencies. We provide our (unoptimized) code in the supplement. 

We apply PRIDA to a set of large, color images that have been synthetically blurred. A recent comparative study on modern blind deconvolution algorithms compiled a dataset of synthetically-blurred spanning a wide range of image sizes, image content, and blur difficulty \cite{lai2016comparative}. 25 real-world images collected from the Internet were each uniformly blurred with 4 known kernels of various size and support. Applying our algorithm to these images we find results comparable to state-of-the-art.

In order to find an appropriate regularization, we perform a mild parameter sweep across all 25 images simultaneously for a given kernel size. For a kernel size of $31 \times 31$, we find that $\lambda = 2e^{-4}$ leads to the best qualitative results. Results on the front page include samples from this set.

To demonstrate the robustness of PRIDA on color images, noise was added to each pixel's lightness value in LAB space \cite{wyszecki1982color} and converted back to the original RGB color space. Figure \ref{fig:colornoise} shows how our recovery is affected by increasing amounts of Gaussian random noise. While our kernel recovery degrades with more added noise, it is clear that we are still able to recover the kernel structure, and that our final recovered image is in fact deblurred. Here, we present the raw output of our proposed model. Since the literature on denoising algorithms is mature, if necessary,
	a denoising algorithm can easily be run after PRIDA to remove the noise depending on its type. In fact, it is a common practice to have a ``non-blind'' stage at the end of the fine scale in many existing deblurring algorithms. 

%% file: conclusions.tex
\section{Conclusion}
We propose a new algorithm, PRIDA, for recovering sharp images through blind deconvolution. PRIDA uniquely takes advantage of the specific problem domain, employing mirror descent over the simplex constraint set. We present theoretical  analysis of PRIDA and derive guarantees on both convergence and robustness with no extra assumptions. In most real world settings, as noted by \cite{zhu2011restoration}, low light conditions and auto-focus software systems may introduce extra blur and noise since they depend on both exposure time and camera settings.

Our exhaustive experimentation shows that PRIDA can be a comprehensive solution for real world problems. We showed  both qualitatively and quantitively that PRIDA performs as good as the state of the art under no noise conditions and unarguably better in the presence of noise.  We believe that our results will be a strong foundation not only for single image blind deconvolution problems, but also for furthering the success of recent data driven approaches such as deep learning architectures. \\

{\bf Our code and additional experiments} can be accessed through our Github repository at \url{https://github.com/sravi-uwmadison/prida}. 


%% file: appendix.tex
\section*{Appendix}

\section{Details of Lemma \ref{triangle_lemma}}
\begin{lemma}
	Let $x^*=\arg\min_{x\in\Delta} \langle z,x\rangle+\KL(x||x^0)$. Then for any $y\in \Delta,z\in\R^s$, we have that, \begin{align}
	\langle z,y\rangle+ \KL(y||x^0)\geq \langle z,x^*\rangle +\KL(x^*||x^0) + \KL(y||x^*).
	\end{align}
	\begin{proof}Define $G(x):=\langle z,x\rangle+\KL(x||x^0)$. Since $x^*$ minimizes $G$ over $\Delta$, and that $G(x)$ is differentiable and strongly convex on $\Delta$ with respect to $\ell_1$-norm, (see page 88 in \cite{Tsybakov:2008:INE:1522486}), the gradient $g$ at $x^*$ should satisfy the following inequality,\begin{align}
		\langle g,x-x^*\rangle\geq 0~\forall~x\in\Delta.\label{eq_1}
		\end{align}		
		Now the derivative of $KL$ divergence with respect to the $i-$th coordinate of $x$ is given by,\begin{align}
		\frac{\partial KL(x,x^0)}{\partial x_i} = 1 + \log \frac{x_i}{x^0_i}	= 1 + \log x_i - \log x_i^0	
		\end{align}
		Plugging in the derivative of KL divergence, adding and
		subtracting $\sum_{i} x^*_i \log x^*_i + y_i \log y_i + x^0_i \log x^0_i$ into \eqref{eq_1},		we get, 
		\begin{align}
		0\leq \langle g,y-x^*\rangle &= \sum_{i=1}^s \left(z_i + 1 + \log x^*_i - \log x_i^0 \right)\left(y_i-x^*_i\right) \label{eq_2}\\
		&=\sum_{i=1}^s \left(z_i\left(y_i-x_i^*\right)\right) + \sum_{i=1}^s y_i- \sum_{i=1}^sx_i^* + \sum_{i=1}^s \left(\log x_i^*-\log x_i^0\right)\left(y_i-x_i^*\right)\label{eq_3}\\
		&=z^T(y-x^*)+ \sum_{i=1}^s \left(\log x_i^*-\log x_i^0\right)\left(y_i-x_i^*\right)\label{eq_4}\\
		&=z^T(y-x^*) +\sum_{i=1}^s\left( y_i\log x_i^* - y_i\log x_i^0 -x_i^*\log x_i^*+x_i^*\log x_i^0\right)\label{eq_5}\\
		&=z^T(y-x^*) +\sum_{i=1}^s \left(y_i\log x_i^* - y_i\log x_i^0\right)- \sum_{i=1}^s \left(x_i^*\log x_i^*-x_i^*\log x_i^0\right)\label{eq_6}\\
		&=z^T(y-x^*) +\sum_{i=1}^s \left(y_i\log x_i^* - y_i\log x_i^0\right) -\KL (x^*||x^0)\label{eq_7}\\
		&=z^T(y-x^*)-\KL (x^*||x^0)\label{eq_8}\\
		&\quad \quad +\sum_{i=1}^s\left(y_i\log x_i^*-y_i\log y_i +y_i\log y_i - y_i\log x_i^0 \right)\nonumber\\
		&=z^T(y-x^*)-\KL (x^*||x^0) -\KL(y||x^*) +\KL(y||x^0)\label{eq_9}
		\end{align}	
		where \eqref{eq_4} is because $y\in \Delta,x^*\in\Delta$. 
		Now rearranging  terms in \eqref{eq_9} (with $\geq 0$ in \eqref{eq_2}) we get the desired result.
	\end{proof}
\end{lemma}
{\bf Comparison of PGD vs PRIDA:} Even though both PGD and PRIDA achieve the same convergence result for smooth function as said in the main paper, PRIDA is more general since the smoothness assumption can be relaxed for any $p\geq 1$ (instead of the specific $p=2$ as required by PGD). This can be seen from inequalities \eqref{conv_eq_2}-\eqref{conv_eq_4} as,
\begin{align}
h(x)&=\leq \loss(y)+ \nabla\loss(y)^T(x-y) + \frac{L}{2}\|x_f-y_f\|^2_2 + \frac{L}{2}\|x_k-y_k\|_p^2+I_{\Delta}(x)&\\
&\leq \loss(y)+ \nabla\loss(y)^T(x-y) + \frac{L}{2}\|x_f-y_f\|^2_2 + \frac{L}{4}KL(x_k||y_k)+I_{\Delta}(x)=:u(x,y)&.
\end{align}
This is most useful when $p=1$ since it amounts to checking (absolute) maximum entry of the Hessian matrix which is easy to perform. 

Moreover, PRIDA can be implemented in an atomic fashion, that is, each coordinate of $k$ can be updated individually followed by a simple normalization, thus the per iteration complexity is $O(s)$. In contrast, the most efficient algorithms to project onto the probability simplex requires at least $O(s\log s)$,  see Figure 1 in \cite{duchi2008efficient}. While the $\log s$ penalty seems innocuous, these algorithms at the least require sorting (as a subroutine) and hence cannot be easily implemented in GPUs.\\


\hrulefill